\documentclass{article}

\usepackage{microtype}
\usepackage{graphicx}
\usepackage{subfigure}
\usepackage{booktabs} % for professional tables
\usepackage{amsmath}
\usepackage{hyperref}

\usepackage{xspace}
\usepackage{paralist}
\usepackage{booktabs} 
\usepackage{multirow}

\usepackage[accepted]{icml2023}
\newcommand{\ourmodel}{\textsc{CEIL}\xspace}

\icmltitlerunning{Compositional Exemplars for In-context Learning}

\begin{document}

\twocolumn[
\icmltitle{Compositional Exemplars for In-context Learning}




\begin{icmlauthorlist}
\icmlauthor{Jiacheng Ye}{hku,pjlab}
\icmlauthor{Zhiyong Wu}{pjlab}
\icmlauthor{Jiangtao Feng}{pjlab}
\icmlauthor{Tao Yu}{hku}
\icmlauthor{Lingpeng Kong}{hku}
\end{icmlauthorlist}

\icmlaffiliation{pjlab}{Shark-NLP, Shanghai Artificial Intelligence Laboratory}
\icmlaffiliation{hku}{Department of Computer Science, The University of Hong Kong}

\icmlcorrespondingauthor{Jiacheng Ye, Zhiyong Wu}{carsonye@connect.hku.hk, whucs2013wzy@gmail.com}

\icmlkeywords{In-context Learning, Large Language Model}

\vskip 0.3in
]



\printAffiliationsAndNotice{}  % leave blank if no need to mention equal contribution

\begin{abstract} 
Large pretrained language models (LMs) have shown impressive In-Context Learning (ICL) ability, where the model learns to do an unseen task via a prompt consisting of input-output examples as the demonstration, without any parameter updates.
The performance of ICL is highly dominated by the quality of the selected in-context examples. However, previous selection methods are mostly based on simple heuristics, leading to sub-optimal performance. In this work, we formulate in-context example selection as a subset selection problem. We propose \ourmodel (\textbf{C}ompositional \textbf{E}xemplars for \textbf{I}n-context \textbf{L}earning), which is instantiated by Determinantal Point Processes (DPPs) to model the interaction between the given input and in-context examples, and optimized through a carefully-designed contrastive learning objective to obtain preference from LMs.
We validate \ourmodel on 12 classification and generation datasets from 7 distinct NLP tasks, including sentiment analysis, paraphrase detection, natural language inference, commonsense reasoning, open-domain question answering, code generation, and semantic parsing. 
Extensive experiments demonstrate not only the state-of-the-art performance but also the transferability and compositionality of \ourmodel, shedding new light on in-context learning. Our code is released at \href{https://github.com/HKUNLP/icl-ceil}{https://github.com/HKUNLP/icl-ceil}.

\end{abstract}
\section{Introduction}

An important goal of artificial intelligence is to develop models that can generalize to unseen tasks. NLP community made a major step towards this goal by discovering the in-context learning (ICL) capability of large pre-trained language models (LMs; ~\citealt{DBLP:conf/nips/BrownMRSKDNSSAA20}). 
% ICL can perform unseen tasks given only a few demonstrative examples as prompt~\citep{liu2021pre}, without any parameter updating. 
Given a limited number of demonstration examples, in-context learning imitates the human ability to leverage prior knowledge to achieve the best generalization performance.

However, such ability comes along with the robustness issue: ICL is particularly sensitive to the selection of in-context examples, and different arrangements can result in a performance deviation from close to random to near state-of-the-art~\citep{rubin-etal-2022-learning,liu2022makes,wu2022self}.
There have been a number of research attempts over the past two years to select better in-context examples. In particular, one prominent approach is to compare the input with each individual example based on learning-free heuristics~\citep{liu2022makes} or learning-based metrics~\citep{rubin-etal-2022-learning}. 
Despite the improved performance, these methods do not take into account the inter-relationship between in-context examples. For instance, the ignorance of redundancy of in-context examples can result in almost identical examples, providing no additional supervision. 
Searching for a compact set of in-context examples becomes even more urgent as there is a hard limit for the prompt length due to the backbone transformer architecture of LMs. 

In this paper, we propose a general approach, named \ourmodel (\textbf{C}ompositional \textbf{E}xemplars for \textbf{I}n-context \textbf{L}earning). Instead of selecting each in-context example independently, \ourmodel models the joint probability of the entire in-context example set, and thus captures the inter-relationship between in-context examples. 
To model the joint probability of a set given a specific input, we propose a novel model based on the conditional determinantal point process (DPP; \citealt{kulesza2012determinantal}) that learns to select the most diverse yet helpful in-context example set (\S\ref{sec:modeling}). To take into account the quality of a selected subset, a scoring function from a language model is incorporated into the conditional DPP to form a contrastive loss (\S\ref{sec:training}). That way, our algorithm maintains the polynomial time maximum a posteriori (MAP) inference of DPP~\citep{chen2018fast} so that the optimal in-context example subset can be found effectively in the inference stage (\S\ref{sec:inference}). 

We validate our method by conducting extensive experiments on 12 classification and generation datasets from 7 distinct tasks, including sentiment analysis, paraphrase detection, natural language inference, commonsense reasoning, open-domain question answering, code generation, and semantic parsing.
The experiments demonstrate that: 1) \ourmodel substantially surpasses both conventional learning-free and learning-based selection approaches, achieving state-of-the-art in-context learning performance (\S\ref{sec:main}); 2) \ourmodel shows transferability across LMs and datasets, enabling a learning-free efficient application (\S\ref{sec:transfer}); 3) \ourmodel inherently learns to compose different examples, shedding new lights on in-context learning for compositional tasks (\S\ref{sec:composition}); 4) \ourmodel is especially effective when the number of in-context examples is in a small scale (\S\ref{sec:analysis}).

\section{Preliminary}

\subsection{In-context Learning}
In-context learning (ICL) refers to one of the core emergent abilities~\citep{wei2022emergent} that infers new tasks from  context~\citep{DBLP:conf/nips/BrownMRSKDNSSAA20}. We use the terms 'in-weights learning' and 'in-context learning' from prior work on sequence models~\citep{DBLP:conf/nips/BrownMRSKDNSSAA20} to distinguish between gradient-based learning with parameter updates and gradient-free learning from context, respectively. 

Formally, each training instance is first linearized into an input text $\mathbf{x} = (x_1\dots x_{\mid \mathbf{x}\mid })$ and an output text $\mathbf{y} = (y_1\dots y_{\mid \mathbf{y}\mid })$, where for all tokens $x_1\dots x_{\mid \mathbf{x}\mid }, y_1\dots y_{\mid \mathbf{y}\mid } \in \mathcal{V}$ and $\mathcal{V}$ is the vocabulary set of the LM.
Given a new test input text $\mathbf{x}_{test}$, in-context learning defines the generation of output $\mathbf{y}$ as
\begin{align*}
    \mathbf{y}_{\textit{test}} \sim \mathcal{P}_{\textit{LM}}(\mathbf{y}_\textit{test}\mid \underbrace{\mathbf{x}_1,\mathbf{y}_1,\dots, \mathbf{x}_K, \mathbf{y}_K}_\textit{context}, \mathbf{x}_{\textit{test}}),
\end{align*}
where $\sim$ refers to decoding strategies (e.g., greedy decoding and nuclear sampling~\citep{holtzman2019curious}), and each in-context example $e_i=(\mathbf{x}_i, \mathbf{y}_i)$ is sampled from a training set $\mathcal{D}=\{(\mathbf{x}_i, \mathbf{y}_i)\}_{i=1}^{N}$. 
The generation procedure is especially attractive as it eliminates the need for updating the parameters of the language model when encountering a new task, which is often expensive and impractical. 

Notably, the performance of ICL on downstream tasks can vary from almost random to comparable with state-of-the-art systems, depending on the quality of the retrieved in-context examples~\citep{rubin-etal-2022-learning,liu2022makes,wu2022self}. Rather than randomly selecting in-context examples for each test input, previous work model the process with a retriever $\mathcal{P}(e_i\mid \mathbf{x}_{test})$, which is either off-the-shelf~\citep{liu2022makes,wu2022self} or fine-tuned~\citep{rubin-etal-2022-learning}.

\subsection{Determinantal Point Processes}

Determinantal point processes (DPPs) are elegant probabilistic models with the ability to express negative interactions.~\citep{kulesza2012determinantal}. 
% \zy{what is "negative interaction"?}
Formally, a DPP $\mathcal{P}$ is a probability measure for $2^{N}$ item sets, where each set consists of items sampled without replacement from a discrete item set $Z = \{1, 2, \dots, N\}$. 
Given the feature vector $\mathbf{a}$ for each item, DPP calculates an $N \times N$ positive semi-definite (PSD) kernel matrix $\mathbf{L}$, where $\mathbf{L}_{ij}$ = $\operatorname{k}(\mathbf{a}_i, \mathbf{a}_j)$ and $\operatorname{k}(\cdot, \cdot)$ is a kernel function. Then the probability over a subset of items indexed by $S \subseteq Z$ can be defined as
\begin{align}
\label{eq:dpp}
    \mathcal{P}(S)=\frac{\operatorname{det}(\mathbf{L}_S)}{\operatorname{det}(\mathbf{L}+\mathbf{I})},
\end{align}
where $\mathbf{L}_S  \equiv [\mathbf{L}_{ij}]_{i,j \in S}$ denotes the restriction of $\mathbf{L}$ to the entries indexed by elements of $S$, $\operatorname{det}(\cdot)$ denotes the determinant of a matrix, and $\mathbf{I}$ is an identity matrix.  
Note according to the the kernel trick~\citep{scholkopf2002learning}, $\operatorname{k}(\mathbf{a}_i, \mathbf{a}_j)$ can be written as $\phi(\mathbf{a}_i)^{T}\phi(\mathbf{a}_j)$, where $\phi(\cdot)$ is a reproducing kernel feature map. 
Therefore, determinants can be geometrically interpreted as the volume of the parallelepiped formed by the vectors $\{\phi(\mathbf{a}_i)\mid i \in S\}$. As the magnitude of an item’s feature vector increases, so do the probabilities of sets containing that item. Meanwhile, as the similarity between two items increases, the probabilities of sets containing both of them decrease. 

Under the distribution $\mathcal{P}$, although the number of possible realizations of $S$ is exponential in $N$, many types of inference tasks including marginalization, conditioning, sampling and MAP inference can be performed in polynomial time~\citep[\emph{inter alia}]{kulesza2012determinantal,gillenwater2012near,han2017faster,chen2018fast}.

\begin{figure*}[ht]
\centering
\includegraphics[width=6in]{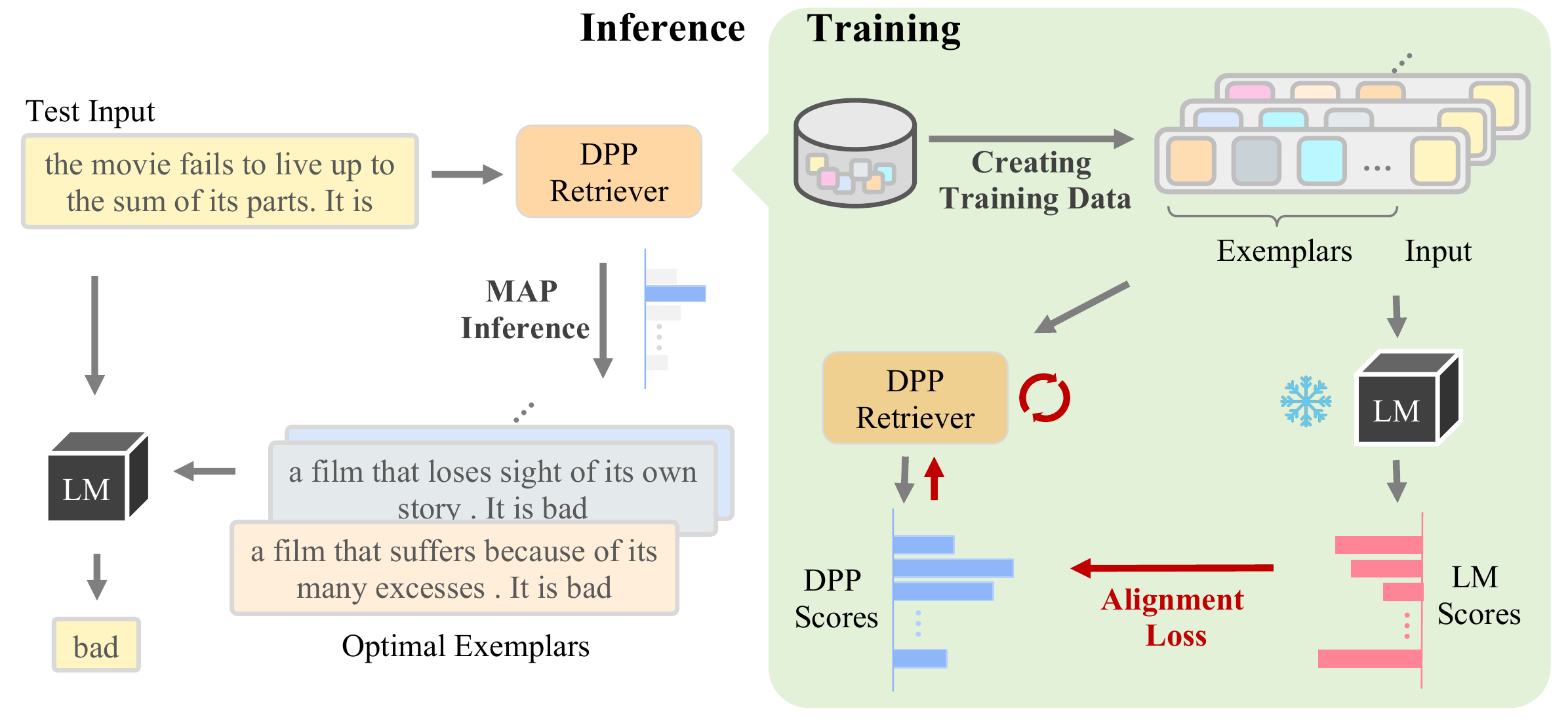}
\caption{\ourmodel at training and inference. Instead of independently retrieving each exemplar (or in-context example), \ourmodel models the entire set of exemplars by learning their joint probability with a conditional DPP (\S\ref{sec:modeling}), which is further trained to align with the LM score through a contrastive loss (\S\ref{sec:training}). For a given test input during inference, the optimal exemplar set is obtained by the learned DPP retriever through MAP inference (\S\ref{sec:inference}). The black-box LM is frozen during the whole procedure.
}
\label{fig:method}
\end{figure*}

\section{Model}
In this section, we introduce an efficient framework, \ourmodel, to learn the \textbf{C}omposition of \textbf{E}xemplars for \textbf{I}n-context \textbf{L}earning, as shown in Figure~\ref{fig:method}.
Instead of independently retrieving each in-context example, \ourmodel models the full in-context example sets by learning the joint probability $\mathcal{P}(S \mid \mathbf{x}_{test})$, and thus captures the inter-relationship between in-context examples.
The joint probability is modeled with a learnable conditional DPP (\S\ref{sec:modeling}) and trained with contrastive learning (\S\ref{sec:training}). In the inference stage, the best in-context example subset is selected via efficient MAP inference (\S\ref{sec:inference}).

% \zy{problem formulation tba}

\subsection{Modeling}
\label{sec:modeling}
For in-context learning, both relevance (i.e., choosing in-context examples similar to the test input) and diversity (i.e., the similarity between examples) are essential, while the vanilla DPPs ignore the relevance term. To infuse both relevance and diversity into the selection procedure, we define a new kernel
\begin{align}
\label{eq:kernal}
\tilde{\operatorname{k}}\left(\mathbf{a}_i, \mathbf{a}_j \mid \mathbf{x}\right)
% \mathbf{\tilde{L}}_{ij}
=\operatorname{g}\left(\mathbf{a}_i, \mathbf{x}\right) \operatorname{k}\left(\mathbf{a}_i, \mathbf{a}_j\right) \operatorname{g}\left(\mathbf{a}_j, \mathbf{x}\right),
\end{align}
which is conditioned on the test input $\mathbf{x}$. The new DPP corresponds to a conditional kernel matrix considering both diversity and relevance: $\mathbf{\tilde{L}}=\operatorname{Diag}(\mathbf{r}) \cdot \mathbf{{L}} \cdot \operatorname{Diag}(\mathbf{r})$, where $\mathbf{r}_{i}=\operatorname{g}\left(\mathbf{a}_i, \mathbf{x}\right)$ is the relevance score for item $i$. 
Based on Eq.~(\ref{eq:dpp}) and Eq.~(\ref{eq:kernal}), we can derive the unnormalized log-probability for subset $S$ as 
% \zy{the derivation of eq 4 seems not that straightforward, maybe add the derivation details in appendix}
\begin{align*}
\log \operatorname{det}\left(\mathbf{\tilde{L}}_{S}\right)=\sum_{i \in S} \log \left(\mathbf{r}_{i}^2\right)+\log \operatorname{det}\left(\mathbf{L}_{S}\right), 
\end{align*}
which clearly shows how the DPP model incorporates the relevance (i.e., $\mathbf{r}_{i}$) and diversity (i.e., $\operatorname{det}(\mathbf{L}_{S})$) of the in-context examples. 

Intuitively, different tasks may prefer a different trade-off between diversity and relevance, e.g., a more complex input may require a more complicated composition of in-context examples. At the same time, the original DPP model does not offer such a mechanism.
To balance the magnitude of diversity and relevance for different tasks, we further incorporate a trade-off parameter $\lambda$ as follows:
\begin{align*}
\log \operatorname{det}\left(\mathbf{L}^{\prime}_{S}\right)=\frac{1}{\lambda} \sum_{i \in S} \mathbf{r}_{i}+ \log \operatorname{det}\left(\mathbf{L}_{S}\right).
\end{align*}
This exactly corresponds to a DPP with kernel $\mathbf{L}^{\prime}=\operatorname{Diag}\left(\exp \left( \frac{\mathbf{r}}{2\lambda}\right)\right) \cdot \mathbf{{L}} \cdot \operatorname{Diag}\left(\exp \left(\frac{\mathbf{r}}{2\lambda}\right)\right)$. 

In practice, the retriever model consists of two embedders to encode input text and in-context examples to their representations $\mathbf{x}$ and $\mathbf{a}$. 
We set both of the two embedders as highly expressive learnable neural networks (e.g., BERT~\citep{devlin-etal-2019-bert}) such that the resulting DPP score (Eq.~(\ref{eq:dpp})) can be an effective ranking metric for subset retrieval. 
On the high-dimensional embedding space, linear kernel (i.e., dot product) is then applied as similarity function $\operatorname{g}$ and $\operatorname{k}$.
The learning of the embedder networks essentially becomes a \textit{metric learning} problem~\citep{kulis2013metric}, which we will introduce in the subsequent section.

\subsection{Training}
\label{sec:training}
Since there is no ground-truth subset of in-context examples for each training instance, we cannot apply the conventional likelihood-maximization method to learn the parameters. In this section, we introduce a contrastive learning framework, with the main idea of rectifying the embedding of each in-context example and training instance such that a `better' subset has a higher probability to be retrieved than a `worse' subset for the training instance.

\paragraph{Training Data.}
Our goal in construction training data is to obtain a dataset $\mathcal{D}_{\textit{train}}=\{(e_i, \{S_{ij}, s_{ij}\})_{j=1}^{M}\}_{i=1}^{N}$ consists of $N$ instances. Each instance contains one input instance $e_i$ from the training set $\mathcal{D}$, $M$ in-context example subsets where each example in subset $S_{ij}$ is also retrieved from $\mathcal{D}$\footnote{We omit the retrieved example that is exactly same as input instance $e_i$ to prevent copying answer.}, and score $s_{ij}$ to indicate the quality of each subset. 

Modeling on the full space of $S$ is exponential in $N$ and thus prohibitive. To this end, we employ a two-stage framework which is commonly used in retrieval~\citep{liu2009learning}. 
We first precompute a set of relevant examples of size $n$ ($n << N$) with a retriever.
Then, we perform non-replacement random sampling to obtain $M$ distinct subsets, with no repeating examples in each subset to prevent zero determinant when calculating $\det(S)$. 

Once we retrieve the set of in-context example subsets $\{S_{ij}\}_{j=1}^{M}$ for each input instance $e_i=(\mathbf{x}_i, \mathbf{y}_i)$, we use the inference LM themselves as the scoring function. To measure the quality of each subset, the score is defined as the probability to predict the answer under the LM, which is formally represented as
\begin{align*}
s_{ij}=\mathcal{P}_{\textit{LM}}\left(\mathbf{y}_i \mid S_{ij}, \mathbf{x}_i\right).
\end{align*}
This indicates how helpful this subset is for decoding the target answer.

\paragraph{Contrastive Loss.}
% We observe it is more challenging to learn which subset of in-context examples is superior than to learn which specific in-context example is superior.
The InfoNCE loss~\citep{oord2018representation} has been found effective to learn which single item is superior to others in various representation learning scenarios~\citep{karpukhin2020dense, he2020momentum, rubin-etal-2022-learning}. However, it has the same treatment for all negative samples and the predicted scores $s_{ij}$ are not fully utilized.
To mitigate this problem, we propose to employ a fine-grained pair-wise margin loss to determine which subset is preferable, and the loss for each training instance is defined as
\begin{align*}
\mathcal{L}_i&=\sum_{\left(S^{+}, S^{-}\right) \in \mathcal{C}_i} \max \left\{0, \frac{\operatorname{log}\mathcal{P}(S^-)- \operatorname{log}\mathcal{P}(S^+)}{c_i}+\xi\right\} \\
c_i&=\max_{S \in \mathcal{C}_i}\log\mathcal{P}(S)-\min_{S \in \mathcal{C}_i}\log\mathcal{P}(S),
\end{align*}
where $\mathcal{C}_i=\{S_{ij}\}_{j=1}^{M}$ contains all the sampled subsets for instance $i$, $\xi$ is set to $\gamma \ast (\operatorname{rank}(S^{-}) - \operatorname{rank}(S^{+}))$ following~\citep{zhong-etal-2020-extractive,an2022cont} to reflect the quality difference in these pairs, $\gamma$ is a hyper-parameter controlling the strength which we set $\gamma=1/|\mathcal{C}_i|$ such that $\xi \in [0, 1]$, and $c_i$ is used to align the scale with $\xi$. 
Note the normalization term $\operatorname{det}(\mathbf{L}+\mathbf{I})$ in Eq.~(\ref{eq:dpp}) requires calculation with complexity $O(N^3)$ on full items with size $N$, while the use of pair-wise ranking loss naturally eliminates the calculation of this term (i.e., 
$\operatorname{log}\mathcal{P}(S^-)- \operatorname{log}\mathcal{P}(S^+)=\log \operatorname{det}\left(\mathbf{L}_{S^-}\right)- \log \operatorname{det}\left(\mathbf{L}_{S^+}\right)$), and thus cuts down the calculation cost.

\subsection{Inference}
\label{sec:inference}
In the inference stage, rather than searching for the most relevant top-k in-context examples as in previous work~\citep{rubin-etal-2022-learning,liu2022makes}, we perform maximum a posteriori (MAP) inference with the learned DPP module, considering both diversity and relevance. The MAP inference of a DPP is defined as 
\begin{align*}
S_{\textit{map}}=\arg \max _{S \subseteq Z} \operatorname{det}\left(\mathbf{L}^{\prime}_S\right),
\end{align*}
which is NP-hard~\citep{ko1995exact}. Similar as in constructing training data, we narrow down the candidate space with KNN retriever from $N$ to $n$. Then we follow \citet{chen2018fast} to use an exact implementation of the greedy algorithm with $O(nK^2)$ complexity, where $K=|S_\textit{map}|$ is the number of in-context examples. In each iteration, the example $j$ is greedily selected based on the incremental gain to the log-probability 
\begin{align*}
j=\arg \max _{i \in Z \backslash S_{\textit{map}}} \log \operatorname{det}\left(\mathbf{L}^\prime_{S_{\textit{map}} \cup\{i\}}\right)-\log \operatorname{det}\left(\mathbf{L}^\prime_{S_{\textit{map}}}\right).
\end{align*}
and added to $S_{\textit{map}}$. With Cholesky decomposition, the complexity can be reduced from $O(nK^3)$ down to $O(nK)$ in each iteration by updating the Cholesky factor incrementally. Note that compared with vanilla KNN retrieval which directly retrieves $K$ examples from $N$, the additional inference latency caused by MAP inference is negligible since both $n$ and $K$ here are relatively small numbers (e.g., $n=100$, $K=16$).

\section{Experiments}
We conduct extensive experiments over 12 diverse datasets, spanning 7 distinct tasks, and show a better approach to in-context learning than previously considered.

% \usepackage{multirow}
\begin{table*}[t]
\centering
\caption{All the datasets and tasks used in the experiments. We show the number of training instances after deduplicating. \#ICE refers to the average number of in-context examples for instances in the validation set when using GPT-Neo as LLM.}
\label{tab:datasets}
\vskip 0.15in
\scalebox{0.95}{
\begin{tabular}{lllccc}
\toprule
\textbf{Type} & \textbf{Dataset} & \textbf{Task} & \textbf{\#Train} & \textbf{\#Validation} & \textbf{\#ICE} \\
\midrule
\multirow{6}{*}{{Classification}} 
 & SST-5~\citep{socher2013recursive} & Sentiment Analysis & 8,534 & 1,101 & 40 \\
 & MRPC~\citep{dolan2004unsupervised} & Paraphrase Detection & 3,668 & 408 & 27 \\
 & MNLI~\citep{williams2018broad} & Natural Language Inference & 392,568 & 19,647 & 40 \\
 & QNLI~\citep{wang2018glue} & Natural Language Inference & 104,707 & 5,463 & 27 \\
 & CMSQA~\citep{talmor-etal-2019-commonsenseqa} & Commonsense Reasoning & 9,740 & 1,221 & 50 \\
 & HellaSwag~\citep{zellers2019hellaswag} & Commonsense Reasoning & 52,611 & 20,006 & 50 \\
 \midrule
\multirow{6}{*}{{Generation}}
 & WebQs~\citep{berant-etal-2013-semantic} & Open-Domain QA & 3,778 & 2,032 & 50 \\
 % & WMT14 En\_De~\citep{bojar-EtAl:2014:W14-33} & Translation & 4,434,423 & 2,169 & 19 \\
 % & XSum~\citep{Narayan2018DontGM} & Summarization & 128,776 & 7,272 & 2 \\
 & GeoQuery~\citep{zelle:aaai96} & Code Generation & 404 & 280 & 50 \\
 & NL2Bash~\citep{LinWZE2018:NL2Bash} & Code Generation & 7,441 & 609 & 43 \\
 & Break~\citep{wolfson2020break} & Semantic Parsing & 44,184 & 7,760 & 28 \\
 & MTOP~\citep{li2021mtop} & Semantic Parsing & 15,564 & 2,235 & 41 \\
 & SMCalFlow~\citep{andreas2020task} & Semantic Parsing & 102,491 & 14,751 & 22 \\
 \bottomrule
\end{tabular}}
\end{table*}

\subsection{Datasets and Evaluation}
All the datasets and tasks are listed in Table~\ref{tab:datasets}. These datasets involve
different task formulations, thereby allowing for extensive evaluations of \ourmodel in varying scenarios.
Prompts and examples of each dataset are shown in Appendix~\ref{app:datasets}.

We compare the predicted answers with the ground truth and report Accuracy (Acc.) for all the classification tasks. For generation tasks, we report Exact Match (EM) for WebQs, GeoQuery, NL2Bash, MTOP, and SMCalFlow, LF-EM~\citep{hasson2021question} for Break following~\citep{rubin-etal-2022-learning}, which is an improvement to EM to measure semantically equivalence.
Final results are reported on the validation set as the test set is private for some datasets.

\subsection{Baselines}
Our model \ourmodel is essentially a learning-based retriever for in-context example selection. We consider both learning-free and other learning-based retrievers as baselines:
\begin{compactitem}
\item \textsc{Random}: The retriever that randomly selects in-context examples from the training set without repetition.
\item \textsc{TopK-BM25}: The classical sparse retrieval method BM25~\citep{stephen2009bm25}, which is an extension of TF-IDF. Top-K-scored examples are selected as in-context examples.
\item \textsc{TopK-BERT}: The dense retriever based on BERT embeddings~\citep{devlin-etal-2019-bert}, we adopt \texttt{bert-base-uncased}\footnote{\href{https://huggingface.co/bert-base-uncased}{https://huggingface.co/bert-base-uncased}} which is publically available in Huggingface Transformers~\citep{wolf-etal-2020-transformers}. 
\item \textsc{DPP-BERT}: The DPP retriever directly uses the original BERT embedding as above without fine-tuning, and adopts MAP inference for subset retrieval~\citep{chen2018fast}.
\item \textsc{TopK-Contriever} and \textsc{TopK-SimCSE}: Two better sentence embedding models trained with contrastive learning~\citep{izacard2021towards,gao2021simcse}.
\item \textsc{EPR}: The learning-based dense retriever trained to retrieve a better singleton in-context example~\citep{rubin-etal-2022-learning}, and Top-K most similar examples are selected in the inference stage. We extend it to other tasks beyond semantic parsing in ~\citet{rubin-etal-2022-learning}.
\end{compactitem}

\subsection{Implementation Details}
We mainly use GPT-Neo~\citep{gpt-neo} as LLM, which is a 2.7B-parameter LM trained on The Pile~\citep{gao2020pile}, an 825 GB text corpus constructed from a wide range of high-quality resources. We also consider GPT2-XL~\citep{Radford2019LanguageMA} (1.5B) and Codex~\citep{chen2021evaluating} (175B) in \S\ref{sec:transfer}.
The number of in-context examples is set to 50, and we truncate it based on the maximum context size for different LMs (e.g., 1,024 for GPT2-XL, 2,048 for GPT-Neo, and 8,001\footnote{\href{https://platform.openai.com/docs/models/codex}{https://platform.openai.com/docs/models/codex}} for Codex) on each task. The resulting average number of in-context examples for each task are listed in Table~\ref{tab:datasets}.

We sort exemplars based on their similarities to the input text in ascending order, in accordance with common practices~\citep{rubin-etal-2022-learning, qiu2022evaluating, levy2022diverse}. 
% Empirically, we find putting the most similar exemplar close to the input works better than the reverse. 
During answer generation, all the classification tasks are reframed into multiple choice following~\citep{DBLP:conf/nips/BrownMRSKDNSSAA20}. We provide the context plus an answer option as input to LM, compare the LM likelihood of each option, and choose the one with the maximum likelihood as the answer. 
On tasks that involve multi-label classification, each label is given a semantically meaningful name as an option (e.g. "Positive" or "Negative" rather than 0 or 1 for sentiment analysis), and then treat the task like multiple choice. For generation tasks, we use greedy decoding to generate answers. 

When constructing data for training the retriever, we limit the number of instances to 44,000 following~\citep{rubin-etal-2022-learning} to reduce the scoring cost, and we sample 50 candidate subsets with 16 examples in each subset for each training instance.
We use Adam optimizer~\citep{kingma2014adam} with batch size 128 and learning rate 1e-5, and run training for 30 epochs on two NVIDIA A100 GPUs.
For each task, we search the trade-off factor $\lambda$ in $\{0.01, 0.05, 0.1\}$. To encode each example into embeddings, we concatenate all the texts in an instance except labels (e.g., premise plus hypothesis in NLI tasks) as input to the BERT-based encoder (i.e., BERT-base with 110M learnable parameters). We initialize the encoder with EPR, which we find significantly helps in training \ourmodel (\S\ref{sec:analysis}).

\begin{table*}[t]
\centering
\caption{Main results on various datasets. We show the absolute performance gain over \textsc{EPR} and \textbf{bold} the best results.}
\label{tab:main}
\vskip 0.15in
\scalebox{0.75}{
\begin{tabular}{lccccccccccccc}
\toprule
% \multirow{2}{*}{\textbf{Method}} & \multicolumn{6}{c}{\textbf{Classification}} & \multicolumn{6}{c}{\textbf{Generation}} & \multirow{2}{*}{\textbf{Avg.}}\\
% \cmidrule(lr){2-7}
% \cmidrule(lr){8-7}
\textbf{Method} & \textbf{SST-5} & \textbf{MRPC} & \textbf{QNLI} & \textbf{MNLI} & \multicolumn{1}{c}{\textbf{CMSQA}} & \textbf{HellaSwag} & \multicolumn{1}{c}{\textbf{WebQs}} & \multicolumn{1}{c}{\textbf{GeoQ.}} & \multicolumn{1}{c}{\textbf{NL2Bash}} & \textbf{Break} & \multicolumn{1}{c}{\textbf{MTOP}} & \multicolumn{1}{c}{\textbf{SMCal.}} & \textbf{Avg.} \\
\midrule
\textit{Learning-free} &  &  &  &  &  &  &  &  &  &  &  &  \\
\textsc{Random} & 31.43 & 67.65 & 56.67 & 37.74 & \textbf{42.51} & 41.16 & 4.87 & 33.93 & 34.35 & 1.70 & 7.30 & 8.90 & 30.68\\
\textsc{TopK-BM25} & 36.06 & 69.36 & 62.29 & 40.68 & 36.12 & 42.20 & 16.68 & 62.86 & 58.98 & 26.00 & 52.70 & 46.10 & 45.84\\
\textsc{TopK-Contriever} & 37.06 & 67.89 & 60.97 & 45.28 & 36.12 & 41.60 & 17.62 & 68.93 & 53.69 & 26.34 & 49.84 & 43.44 & 45.73 \\
\textsc{TopK-SimCSE} & 37.06 & 66.91 & 61.58 & 44.85 & 35.54 & 41.69 & 16.83 & 66.43 & 54.89 & 26.58 & 47.29 & 42.59 & 45.19 \\
\textsc{TopK-BERT} & 37.24 & 69.36 & 64.65 & 42.15 & 35.38 & 40.28 & 17.08 & 66.79 & 51.30 & 26.84 & 52.13 & 44.63 & 45.65 \\
\textsc{DPP-BERT} & 36.78 & 69.61 & 63.83 & 39.60 & 37.26 & 40.69 & 14.57 & 70.71 & 48.99 & 26.70 & 53.14 & 43.26 & 45.43\\
\textit{Learning-based} &  &  &  &  &  &  &  &  &  &  &  &  \\
\textsc{EPR} & 42.82 & 75.98 & 80.76 & 66.06 & 36.77 & 42.61 & 19.59 & 68.57 & 56.82 & 31.90 & 64.20 & 54.30 & 53.37\\
\ourmodel & \textbf{47.05} & \textbf{80.15} & \textbf{85.41} & \textbf{71.74} & {37.18} & \textbf{43.20} & \textbf{20.92} & \textbf{73.21} & \textbf{59.91} & \textbf{34.18} & \textbf{67.43} & \textbf{60.73} & \textbf{56.76}\\
$\Delta$ Absolute gain & {+4.23} & {+4.17} & {+4.65} & {+5.68} & {+0.41} & {+0.59} & {+1.33} & {+4.64} & {+3.09} & {+2.28} & {+3.23} & {+6.43} & {+3.39}\\
\bottomrule
\end{tabular}}
\end{table*}

\begin{table*}
\centering
\caption{Results on compositional semantic parsing datasets using GPT-Neo and Codex as inferencers. The retriever used for Codex is the same as that for GPT-Neo, and is trained on the GeoQuery and SMCalFlow datasets. 0-S referring to a non-compositional test set and $k$-C referring to a compositional test set with additional $k$-shot compositional examples as demonstrations ($k\in\{0,8,16,32\}$; see Appendix~\ref{app:com-exp} for details). We show the absolute performance gain over \textsc{EPR} and \textbf{bold} the best results.}
\label{tab:composional}
\vskip 0.15in
\scalebox{0.9}{
\begin{tabular}{lccccccccc}
\toprule
\multirow{2}{*}{{\textbf{Model}}} & \multicolumn{4}{c}{\textbf{GeoQuery}} & \multicolumn{5}{c}{\textbf{SMCalFlow-CS}} \\
\cmidrule(lr){2-5}
\cmidrule(lr){6-10}
 & Standard & Template & TMCD & Length & 0-S & 0-C & 8-C & 16-C & 32-C \\ 
 \midrule
 \textit{Previous Results} &  &  &  &  &  &  &  &  &  \\
 {T5 Base + CSL-Aug}~\citep{qiu-etal-2022-improving} & 93.30 & 89.30 & 74.90 & 67.80 & \multicolumn{5}{c}{\multirow{2}{*}{(Different Dataset Version)}} \\
{Cover-LS}~\citep{levy2022diverse} & 91.40 & 81.60 & 76.30 & 70.00 & \multicolumn{5}{c}{} \\
PaLM 540B~\citep{qiu2022evaluating} & 86.80 & 76.60 & 63.60 & 57.90 & {-} & {-} & 4.70 & 5.00 & 11.70 \\
PaLM 540B (Oracle)~\citep{qiu2022evaluating} & 92.10 & 77.93 & 73.83 & 63.90 & - & - & 33.90 & 36.70 & 45.60 \\
\midrule
\textit{GPT-Neo 2.7B Inferencer} &  &  &  &  &  &  &  &  &  \\ 
\textsc{TopK-BERT} & 66.79 & 30.75 & 41.82 & 31.59 & 31.94 & 0.00 & 0.28 & - & - \\ 
\textsc{EPR} & 68.57 & 38.95 & 44.09 & 32.27 & 57.78 & 0.00 & 0.00 & - & - \\ 
\ourmodel & \textbf{73.21} & \textbf{40.77} & \textbf{44.09} & \textbf{32.73} & \textbf{60.27} & 0.00 & \textbf{0.28} & - & - \\ 
$\Delta$ Absolute gain & {+4.64} & {+1.82} & {+0.00} & {+0.46} & {+2.49} & {+0.00} & {+0.28} & {-} & {-} \\
\midrule
\textit{Codex 175B Inferencer} &  &  &  &  &  &  &  &  &  \\
\textsc{TopK-BERT} & 91.79 & 87.47 & 61.36 & 69.55 & 80.83 & 0.00 & 40.83 & 46.67 & 49.72 \\ 
\textsc{EPR} & 91.70 & 87.93 & 62.73 & 73.41 & 80.83 & 0.56 & 35.56 & 38.61 & 48.06 \\ 
\ourmodel & \textbf{93.21} & \textbf{89.98} & \textbf{63.64} & \textbf{74.09} & \textbf{81.39} & \textbf{1.67} & \textbf{42.78} & \textbf{48.06} & \textbf{55.28} \\
$\Delta$ Absolute gain & {+1.51} & {+2.50} & {+0.91} & {+0.68} & {+0.56} &{+1.11} & {+7.22} & {+9.45} & {+7.22} \\
\bottomrule
\end{tabular}}
\end{table*}

\subsection{Main Results}
\label{sec:main}

We experiment on 12 datasets spanning 7 distinct tasks and the results are shown in Table~\ref{tab:main}. 
Overall, we found generation tasks benefit more from a better set of in-context examples than classification tasks. For example, the simple \textsc{TopK-BM25} retriever brings an around 12\% to 45\% absolute performance gain compared to the \textsc{Random} retriever. The underlying reason can be that relevant answers rarely appear in the non-relevant exemplars for the generation tasks.

We find \ourmodel substantially outperforms learning-free baselines and is especially effective
on Natural Language Inference (NLI) tasks (e.g., QNLI, MNLI), where more than 20\% absolute improvements are obtained. On most of the other classification and generation tasks, \ourmodel surpasses the learning-free retrievers by around 10\%, with an exception on Commonsense Reasoning tasks (i.e., CMSQA and HellaSwag). Interestingly, all the other retrievers (e.g., \textsc{TopK-BM25}, \textsc{TopK-BERT} and \textsc{EPR}) perform comparable to the random retriever on this task, indicating the related commonsense knowledge may not exists in the training data.

Compared with the learning-based retriever, \ourmodel consistently outperforms \textsc{EPR} on all the tasks, suggesting the effectiveness of bringing interaction between in-context examples into the learning procedure. Note \ourmodel introduces no additional parameters compared with \textsc{EPR} and the learning-free \textsc{TopK-BERT}, suggesting \ourmodel is not only effective but also can be efficiently applied in real applications with no deployment cost.

\subsection{Compositionality}
\label{sec:composition}

A natural intuition of the superior performance of \ourmodel is that it learns to compose exemplars such that the whole subset helps in predicting answers. To systematically investigate the compositional ability of the learned retriever, we experiment on two well-designed semantic parsing datasets obtained from original SMCalFlow and GeoQuery datasets, where the test examples requires explicit compositional exemplars (e.g., to predict the program of ``organize an event with my manager'', one has to retrieve exemplars relates to ``organize an event'' and ``my manager''). 
We evaluate the trained retrievers in \S\ref{sec:main} on various data splits in these two datasets (see Appendix~\ref{app:com-exp} for details), and the results are shown in Table~\ref{tab:composional}.

The \texttt{Template} and \texttt{Standard} splits account for the majority of the performance difference between \ourmodel and \textsc{EPR}, with around 2\% and 5\% on GeoQuery dataset. 
Meanwhile, the improvement on all the cross-domain splits ($k$-C)
% \zy{what is k and c}
of SMCalFlow-CS excel the single-domain split (0-S) when comparing \ourmodel with \textsc{TopK-BERT} and \textsc{EPR}. 
These indicate \ourmodel does, to a certain extent, retrieve compositional exemplars. 
Overall, \ourmodel improves performance on all the difficult splits on these two datasets, indicating better organizing the in-context examples helps in predicting compositional and longer target programs.

The previous solutions to generating compositional programs require compositional data augmentation for training LM~\citep{qiu-etal-2022-improving}, or test-time local-structure prediction for selecting diverse exemplars~\citep{levy2022diverse}. \ourmodel can be seen as an alternative approach that directly retrieves a diverse exemplars subset without tuning inference LM, which is expensive, or test-time question decomposition, which impairs efficiency and may suffer from error propagation. Note though the inference LM in \ourmodel hasn't seen any compositional data in the context, the retriever has seen as it needs to be trained in the standard dataset. 
An interesting further work would be training a retriever that directly generalizes to unseen compositional tasks without seeing any compositional data, as we have shown the possibility of transferring across datasets in \S\ref{sec:transfer}.

\subsection{Transferability}
\label{sec:transfer}
The compositional characteristics of natural language are general, meaning the retriever may exploit similar knowledge in different tasks or inference LMs. 
This motivates us to explore whether the retriever trained on one dataset and LM inferencer can be directly transferred to others without further tuning. 
This is a practical research question as training a retriever for each dataset or LM inferencer can be costly in real applications.

\begin{figure}[t]
\centering
\includegraphics[width=3.2in]{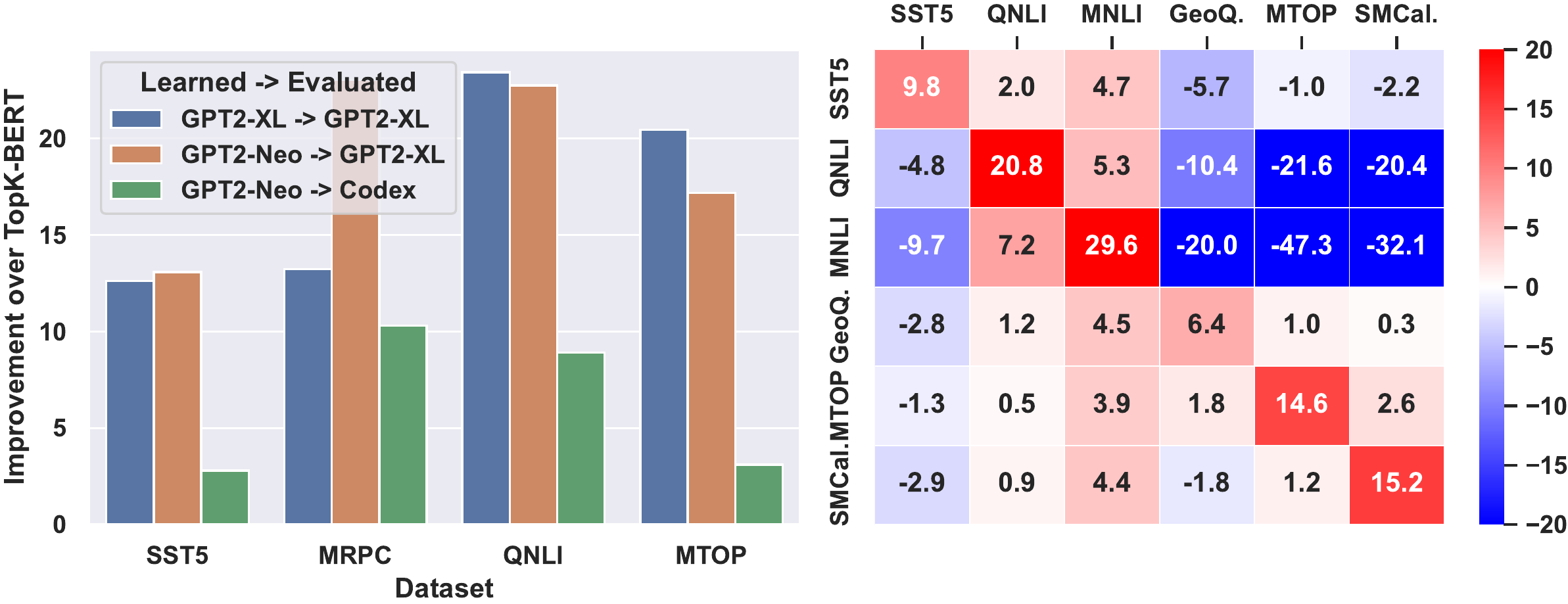}
\caption{\textbf{(Left)} Results of transferring a retriever learned on one LM inferencer to others. \textbf{(Right)} Results of transferring a retriever learned on one dataset (row) to others (column). For both figures, we show the absolute improvement over \textsc{TopK-BERT}. 
% See Appendix\jc{todo} for detailed results.
}
\label{fig:transfer}
\end{figure}

\paragraph{Transfer across LMs}
We consider transferring the retriever trained on GPT-Neo to a similar-sized model GPT2-XL~\citep{Radford2019LanguageMA} (1.5B) and a much larger model Codex~\citep{chen2021evaluating} (175B). Note in the transfer setting, \ourmodel becomes a learning-free method under the target LM, thus we also compare the results with \textsc{TopK-BERT}. We show the absolute improvement over \textsc{TopK-BERT} in Figure~\ref{fig:transfer} (Left). 
Interestingly, the retriever learned on GPT2-Neo performs comparably with that on GPT2-XL when evaluating on GPT2-XL for datasets such as SST5, QNLI, and MTOP. We also surprisingly find the transferred retriever outperforms the specially-trained one on the MRPC dataset, indicating it may bring extra knowledge (e.g., compositional characteristic of natural language) beyond learning from the target LM.
Note when considering a large LM (e.g., Codex) as the LM inferencer, learning an LM-specific retriever can be costly due to the restricted access.
Though \textsc{TopK-BERT} already performs well on Codex, \ourmodel still brings improvement.

\paragraph{Transfer across Datasets}
We further investigate whether a retriever trained on one dataset transfers to others, as shown in Figure~\ref{fig:transfer} (Right). We find almost all the retrievers transfer to NLI tasks such as QNLI and MNLI, and achieve better performance than \textsc{TopK-BERT}. However, the NLI-trained retrievers hardly transfer to other tasks except for NLI task (e.g., QNLI-trained retriever only benefits MNLI). We conjecture that this is due to the fact that NLI tasks require two text inputs, but other tasks only require one, and that knowledge gained from single-input tasks still has value in double-input tasks. For other single input tasks, we find only the retriever learned on similar tasks (e.g., Code Generation and Semantic Parsing) shows transferability. Developing a retriever works for all tasks is a challenging but valuable research topic, which we leave for future work.

\subsection{Analysis}
\label{sec:analysis}

\paragraph{On the Effect of Training Data}
To investigate the effect of training data, we compare different candidate sampling strategies and the number of candidates. Beyond sampling candidates randomly, we also sample fix-sized candidates based on probability defined by k-DPP~\citep{kulesza2011k}. We always include the Top-K candidate, thus we also report $MRR=\frac{1}{N}\sum_{i=1}^N \frac{1}{rank_i}$ to measure the quality of the training data based on the ranking of the Top-K candidate among all the candidates. A lower MRR means that there are more candidates that are "better" than the Top-K. 
As shown in Table~\ref{tab:ana-data}, the one-stage random retrieval greatly degrades performance on SST5 and MTOP datasets. Surprisingly, the MRR of one-stage random retrieval achieves the lowest, indicating relevance is not the only factor that contributes to the quality of a subset. 
Two-stage random sampling slightly outperforms k-DPP sampling with similar MRR. 
Furthermore, we find the number of candidates mostly affects generation tasks, which is considered to be more complex than classification and increasing the number improves the final performance.

\begin{table}[t]
\centering
\caption{Results of various sampling strategies and number of candidates (C) per instance in construction training data. We report both MRR of the Top-K candidate and the performance of the trained retriever.}
\label{tab:ana-data}
\vskip 0.15in
\scalebox{0.68}{
\begin{tabular}{lcccc}
\toprule
{\textbf{Method}} & \textbf{SST5} & \textbf{MRPC} & \textbf{GeoQuery} & \textbf{MTOP} \\
\midrule
% \textsc{EPR} & 42.82 & 75.98 & 68.57 & 64.20 \\
\textsc{Rand, C50} & 0.08/35.97 & 0.08/80.88 & 0.08/71.07	& 0.07/56.60 \\
\textsc{Top100+Rand, C10} & 0.29/46.14 & 0.29/\textbf{81.37} & 0.27/67.86 & 0.25/62.37 \\
\textsc{Top100+Rand, C50} & 0.09/\textbf{47.05} & 0.09/80.15  & 0.08/\textbf{73.21} & 0.09/\textbf{67.43} \\
\textsc{Top100+k-DPP, C50} & 0.09/45.96 & 0.09/79.41  & 0.09/71.07 & 0.09/63.62 \\
\bottomrule
\end{tabular}}
\end{table}

\begin{table}[t]
\centering
\caption{Comparisons of different initializations and contrastive losses for \ourmodel.}
\label{tab:loss}
\vskip 0.15in
\scalebox{0.68}{
\begin{tabular}{lccccc}
\toprule
{\textbf{Method}} & \textbf{SST5} & \textbf{MRPC} & \textbf{QNLI} & \textbf{GeoQuery} & \textbf{MTOP} \\
\midrule
\textit{Baselines} \\
\textsc{TopK-BERT} & 37.24 & 69.36 & 64.65 & 66.79 & 52.13 \\
\textsc{EPR} & 42.82 & 75.98 & 80.76 & 68.57 & {64.20} \\
\textit{Training Strategies} \\
\textsc{BERT init + InfoNCE} & 31.34 & 69.12 & 63.92 & 68.57 & 47.43 \\
\textsc{BERT init + Pair-wise} & {35.55} & 67.89 & 65.00 & {67.50} & {41.30} \\
\textsc{EPR init + InfoNCE} & \textbf{49.14} & \textbf{80.64} & \textbf{85.54} & 69.29 & 61.92 \\
\textsc{EPR init + Pair-wise} & 47.05 & 80.15 & 85.41 & \textbf{73.21} & \textbf{67.43} \\
\bottomrule
\end{tabular}}
\end{table}
\paragraph{On the Effect of Learning Strategies}
We compare different initializations and contrastive losses in Table~\ref{tab:loss}. Learning which subset is superior based on the raw BERT encoders is challenging, but using \textsc{EPR} as an initializer greatly improves performance. This indicates the knowledge learned from a single in-context example selection contributes to the set-level selection. Regarding the choice of contrastive loss, we find InfoNCE and pair-wise margin loss perform comparably on classification tasks, but the latter significantly surpasses the former on generation tasks, with approximately 4\% and 6\% on GeoQuery and MTOP, respectively. Note that generation tasks are more difficult than classification as the answers rarely appear in the in-context examples directly.
This indicates pair-wise margin loss, which is a more fine-grained contrastive loss than InfoNCE loss, better displays its effectiveness on much harder tasks.

\begin{table}[t]
\caption{Comparison of inference algorithms, i.e., \textsc{TopK} and \textsc{DPP} (short for \textsc{DPP-MAP})),  on \textsc{BERT}, \textsc{EPR} and \ourmodel.}
\centering
\label{tab:inference}
\vskip 0.15in
\scalebox{0.68}{
\begin{tabular}{lcccccc}
\toprule
\textbf{Method} & \textbf{SST5} & \textbf{MRPC} & \textbf{MNLI} & \textbf{CMSQA} & \textbf{MTOP} & \textbf{SMCal.} \\
\midrule
\textit{learning-free} &  &  &  &  &  &  \\
\textsc{TopK-BERT} & 37.24 & 69.36 & 42.15 & 35.38 & 52.13 & 44.63 \\
\textsc{DPP-BERT} & 36.78 & 69.61 & 39.60 & 37.26 & 53.14 & 43.26 \\
\textit{learning-based} &  &  &  &  &  &  \\
\textsc{TopK-EPR} & 42.82 & 75.98 & 66.06 & 36.77 & 64.20 & 54.30 \\
\textsc{DPP-EPR} & 45.54 & 80.39 & 65.09 & 35.54 & 64.38 & 57.64 \\
\textsc{TopK-CEIL} & 45.78 & \textbf{81.37} & 71.25 & 37.10 & 66.62 & 59.95 \\
\textsc{DPP-}\ourmodel & \textbf{47.05} & 80.15 & \textbf{71.74} & \textbf{37.18} & \textbf{67.43} & \textbf{60.73} \\
\bottomrule
\end{tabular}}
\end{table}

\paragraph{On the Effect of Inference Strategies}
In this paragraph, we compare two inference algorithm (i.e., \textsc{TopK} and \textsc{DPP} (short for \textsc{DPP-MAP})) across learning-free and learning-based methods.
Compared with \textsc{TopK}, we find \textsc{DPP-MAP} brings more improvement when using a learning-based retriever, indicating the importance of aligning the 'similarity' of embedding to the 'usefulness' for inference.
Beyond accuracy, we also find the latency of retrieving 50 in-context examples for \textsc{TopK} and \textsc{DPP-MAP} on SST5 dataset are 30s and 36s (1.2x), respectively. Thus, we recommend choosing \textsc{TopK} or \textsc{DPP-MAP} for different tasks considering the additional inference cost in real applications.
We provide more details on the performance-efficiency trade-off in Appendix

\begin{figure}[t]
\centering
\includegraphics[width=3.2in]{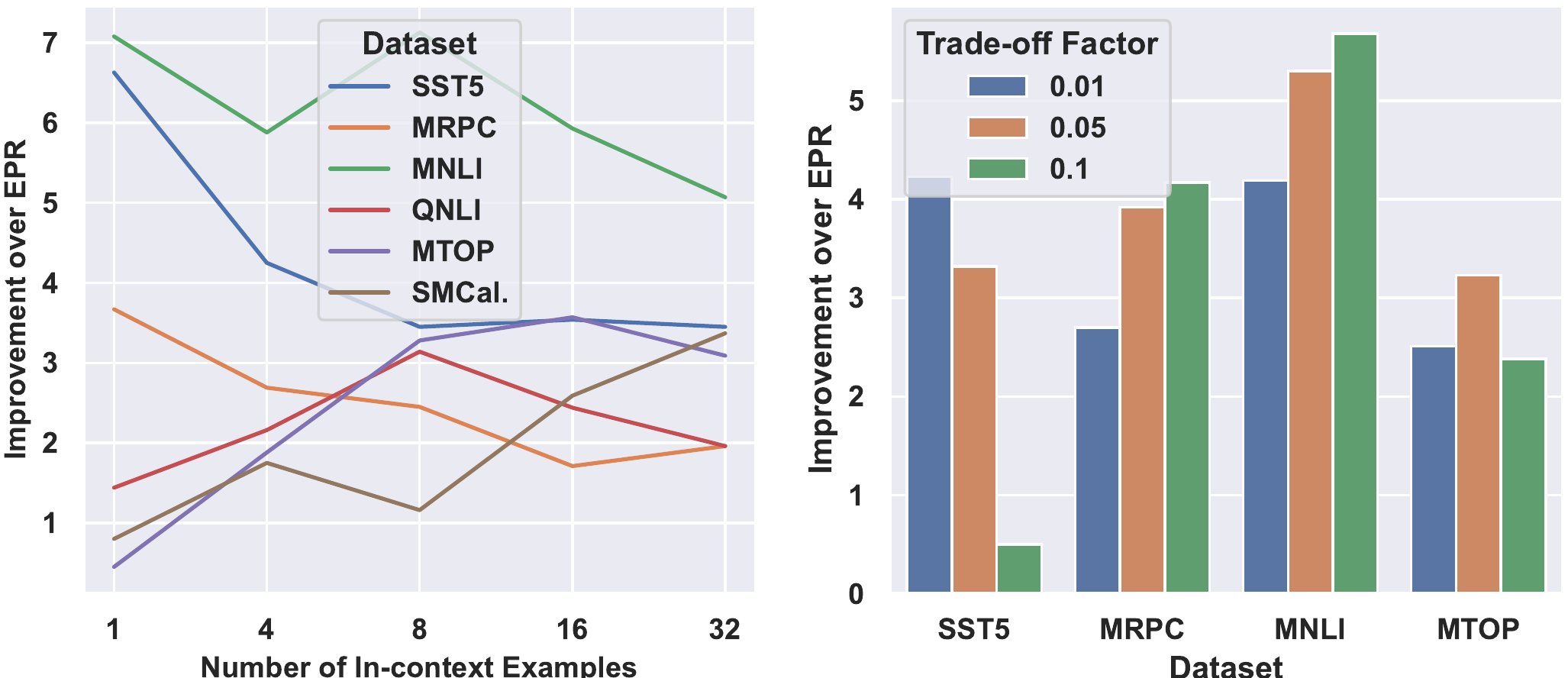}
\caption{\textbf{(Left)} Comparison of different number of in-context examples on various datasets.\textbf{(Right)} Comparison of different trade-off factors on various datasets. For both figures, we show the absolute improvement over \textsc{EPR}.}
\label{fig:ice_trade_off}
\end{figure}

\paragraph{On the Effect of In-context Example Numbers}
Most of the current large LMs are trained with a limited input length such as 1,024 in GPT2-XL and 2,048 in GPT2-Neo, which restricts the maximum number of in-context examples. 
Here we evaluate the trained retriever under various number of in-context examples, as shown in
Figure~\ref{fig:ice_trade_off} (Left).
We find a clear increasing trend for most classification tasks when decreasing the numbers, indicating the effectiveness in selecting a compact set of in-context examples. We observe an opposite trend in generation tasks, which we hypothesize is because the difficulty of generation tasks. i.e., the question can only be answered with a sufficient number of in-context examples. Another advantage of a compact set of in-context examples is we can greatly cut down the computations, as the attention module ~\citep{vaswani2017attention} in most LMs is of quadratic complexity. We find \ourmodel mostly outperforms \textsc{EPR} and \textsc{TopK-BERT} with 32 in-context examples by using merely 4 and 1 example, respectively (see Appendix~\ref{app:num-ice} for details).

\paragraph{On the Effect of Trade-off Factor}
We perform an ablation study to see the effect of trade-off factor in Figure~\ref{fig:ice_trade_off} (Right). Note a smaller factor put more emphasize on the relevance. We find the best performing factor varies for different datasets. A general observation is that diversity is more important for more difficult tasks, such as NLI and semantic parsing, but relevance is more crucial for the simpler tasks such as sentiment analysis. Given the discrepancy, we find introducing the trade-off factor still consistently outperforms \textsc{EPR} baselines that only considers relevance, verifying the effectiveness of \ourmodel.

\section{Related Work}
% Some of them propose heuristic metrics such as surface form or embedding similarity~\citep{liu2022makes}, while others employ LM-based metrics such as mutual information~\citep{sorensen-etal-2022-information} and perplexity~\citep{gonen2022demystifying}.
% Nevertheless, the former suffers from the unsatisfying performance and the latter requires multiple forward passes through LM, which incurs test-time efficiency.
% Large language models have taken NLP research into a new phase with their mind-blowing few-shot performance achieved via in-context learning.  without computationally intensive parameter updating. 
% \subsection{Prompting LMs}

\subsection{In-context Learning}
By providing a few input-output examples as demonstrations, in-context learning (ICL) empowers large language models (LMs) to ``learn by analogy'' and perform complex tasks such as web browsing~\citep{nakano2021webgpt}, coding~\citep{chen2021codex}, data generation~\citep{ye-etal-2022-progen,Ye2023GeneratingDF}, strategic game~\citep{meta2022human}, and conversations~\citep{openai2022chatgpt}. The popularity of ICL also raises growing concerns regarding its instability: given different selections, ICL's performance can vary from near state-of-the-art to random~\citep{liu2022makes}. To mitigate this issue, researchers have made significant efforts on in-context example selection, which can be cataloged into \textit{learning-free} and \textit{learning-based} methods. In the line of learning-free methods, various heuristic criteria are proposed, such as the semantic similarity between testing examples and demonstrations~\citep{liu2022makes}, entropy~\citep{lu2022fantastically,wu2022self}, diversity~\cite{ye2022complementary,su2022selective,levy2022diverse,agrawal2022context}. However, learning-free methods generally require human experts to design task-specific heuristics and lead to sub-optimal performance. Researchers thus have started to explore learning-based methods to push the envelope further. \citet{rubin-etal-2022-learning} propose to train a singleton example scorer using contrastive learning with signals from LM inferencer. In comparison, we aim to jointly model the selection of the entire exemplar set, which additionally considers the interaction between in-context examples. 
Beyond in-context example selection, some works have explored multi-pass ICL, which first generates multiple responses from various subsets of exemplars~\citep{shi2022natural,li2022advance} and then aggregate them through techniques similar to self-consistency~\citep{wang2022self}.
In contrast, multi-pass ICL approaches require multiple test-time inferences, which can result in inefficiency.

\subsection{Determinantal Point Processes}
Determinantal point processes (DPPs) are efficient probabilistic models that can measure both the diversity and quality of items in a subset, which makes it a natural choice for the diverse subset selection problem~\citep{kulesza2012determinantal}. DPPs have been applied for document and video summarization~\citep{kulesza2011k,gong2014diverse}, recommendation systems~\citep{gillenwater2012near}, object detection~\citep{azadi2017learning} and multi-label classification~\citep{xie2017deep}. Most recently, DPPs have been employed in in-context learning specially for compositional tasks~\citep{levy2022diverse}, where the authors first predict all possible target subphrases with a specially-trained model, and then adopt DPPs to sample a diverse subset of in-context examples to cover as many subphrases as possible. However, the diversity objective in DPPs is not aligned with LMs and is generally task-specific. In contrast, we frame DPPs into an end-to-end framework, which not only captures the interaction between in-context examples but also well reflects the preference of LMs on the probability of DPPs.

\section{Conclusion}

In this paper, we recast in-context example selection into an end-to-end optimization problem. We propose \ourmodel, which leverages DPP to model the probability of the entire subset of in-context examples, and is learned through a contrastive learning framework. Results on 7 classification and generation tasks with 12 different benchmarks show that \ourmodel clearly beats previous competitive methods. The learned retriever in \ourmodel also exhibits surprising transferability across LMs and datasets, and compositionality for compositional tasks, showing an effective and efficient approach to adapt the black-box large LMs to the downstream tasks.

\section*{Acknowledgement}
We thank the anonymous reviewers whose suggestions helped clarify this work. This work is partially supported by the Shanghai Committee of Science and Technology (Grant No.~21DZ1100100), and the joint research scheme of the National Natural Science Foundation of China (NSFC) and the Research Grants Council (RGC) under grant number N\_HKU714/21.


\bibliography{main}
\bibliographystyle{icml2023}

\appendix
\section{Experimental Setup}
\label{app:com-exp}

\begin{table*}[htb]
\centering
\caption{Datasets with corresponding prompts and examples used in the experiments.}
\label{tab:examples}
\includegraphics[width=6.5in]{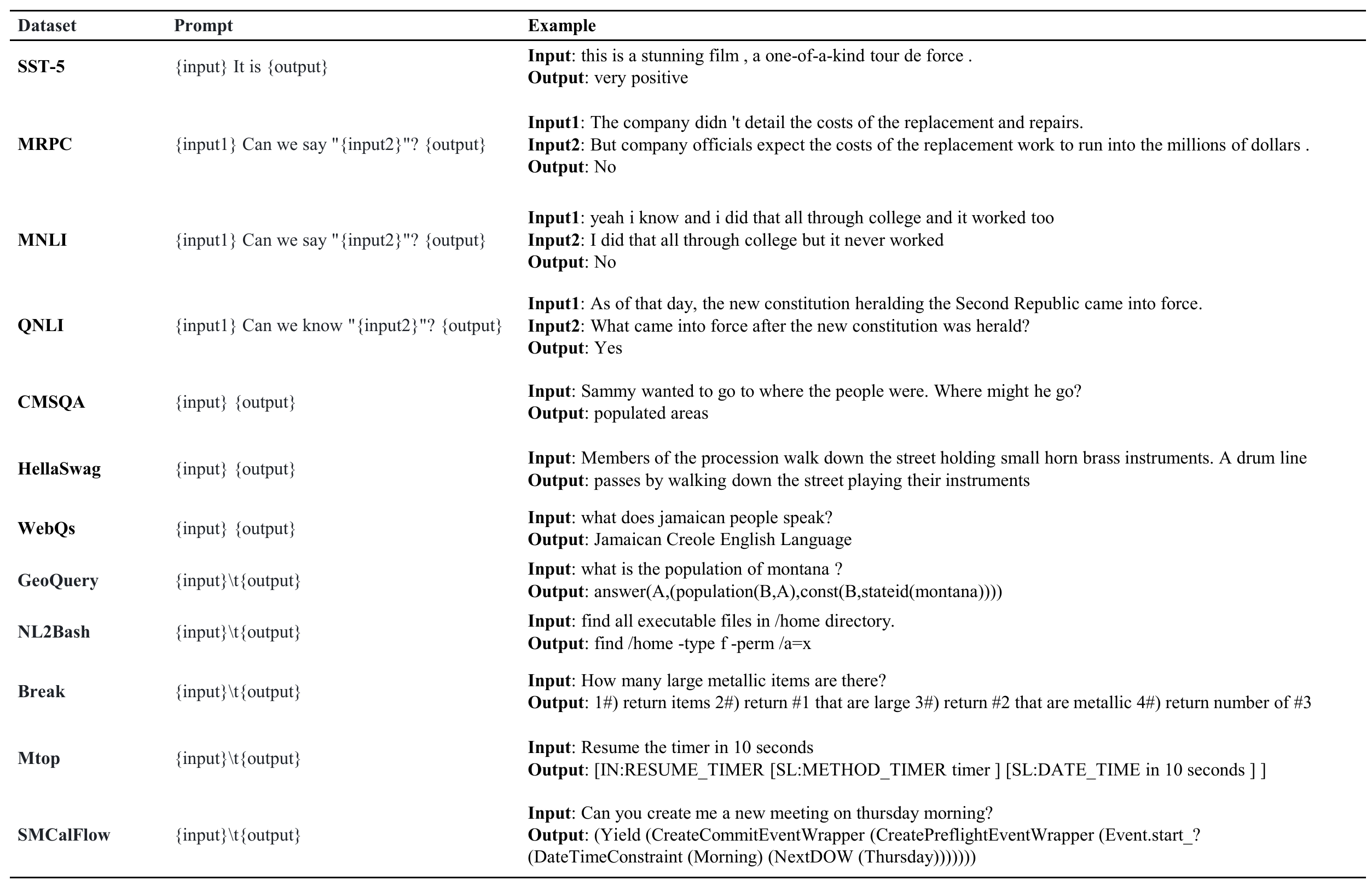}
\end{table*}

\subsection{Datasets} 
\label{app:datasets}
We conduct experiments on 12 classification and generation tasks, and examples in each dataset are shown in Table~\ref{tab:examples}. We illustrate the detail of each dataset as follows. 

\paragraph{SST-5~\citep{socher2013recursive}} is a sentiment classification benchmark containing five fine-grained classes including `very positive', `positive' `neutral', `negative', and `very negative'. 
\paragraph{MRPC~\citep{dolan2004unsupervised}} is a corpus of sentence pairs automatically extracted from online news sources, with human annotations for whether the sentences in the pair are semantically equivalent. 
\paragraph{MNLI~\citep{williams2018broad}} is a crowdsourced collection of sentence pairs with textual entailment annotations. Given a premise sentence and a hypothesis sentence, the task is to predict whether the premise entails the hypothesis (entailment), contradicts the hypothesis (contradiction), or neither (neutral). 
\paragraph{QNLI~\citep{wang2018glue}} is a question-answering dataset consisting of question-paragraph pairs, and the task is to determine whether the context sentence contains the answer to the question.
\paragraph{CMSQA~\citep{talmor-etal-2019-commonsenseqa}} (short for CommonsenseQA) is a multiple-choice question-answering dataset that requires different types of commonsense knowledge. The task is to predict the correct answer out of five provided candidate answers. 

\paragraph{HellaSwag~\citep{zellers2019hellaswag}} is a large-scale dataset of grounded commonsense reasoning.
There are four candidate answers for each question: a video caption from ActivityNet Captions~\citep{heilbron2015activitynet} and the Large Scale Movie Description Challenge~\citep{rohrbach2017movie}. The three incorrect answers are adversarially generated and human validated to deceive machines. The correct answer is the actual video caption for the subsequent occurrence in the video.

\paragraph{WebQs~\citep{berant-etal-2013-semantic}} (short for WebQuestions) is question-answer pairs obtained from the web. The questions are selected using Google Suggest API,
and the answers are entities in Freebase.

\paragraph{NL2Bash~\citep{LinWZE2018:NL2Bash}} is a dataset for the problem of mapping English sentences to Bash commands. The corpus consists of text–command pairs, where each
pair consists of a Bash command scraped from the web and an expert-generated natural language description.

\paragraph{GeoQuery~\citep{zelle:aaai96,shaw-etal-2021-compositional}} contains a parallel
corpus of 880 English questions about US geography paired with Prolog queries.
The compositional dataset of GeoQuery were created by  
 ~\citet{shaw-etal-2021-compositional}, focusing on compositional generalization. In addition to the original \texttt{Standard} split, it contains three additional splits: (1) the \texttt{Template} split, where abstract output templates in training and test data are disjoint~\citep{finegan-dollak-etal-2018-improving}; (2) the \texttt{TMCD} split, which makes the distributions of compounds in training and test data as divergent as possible; and (3) the \texttt{Length} split, where the test instances are longer than the training ones.

\paragraph{Break~\citep{wolfson2020break}} is a dataset that maps complex natural language questions into a language-based meaning representation. The question is decomposed into an ordered list of atomic steps, which is used as the target sequence. We use the low-level Break subset following~\citep{rubin-etal-2022-learning}.

\paragraph{MTOP~\citep{li2021mtop}} is a multilingual task-oriented semantic parsing
dataset covering 6 languages and 11 domains.  The target commands are complex queries featuring nested intent-slot prediction.
Similar to past work ~\citep{rubin-etal-2022-learning}, we use the English subset of MTOP.

\begin{table*}[h]
\centering
\caption{Inference latency on SST-5 validation set and evaluation metrics on different datasets when varying $n$ at inference time.}
\label{tab:vary-n}
\vskip 0.15in
\scalebox{1}{
\begin{tabular}{lc|ccccccc}
\toprule
\textbf{Model} & \textbf{Latency} & \textbf{SST5} & \textbf{MRPC} & \textbf{QNLI} & \textbf{GeoQuery} & \textbf{NL2Bash} & \textbf{MTOP} & \textbf{Avg.} \\
\midrule
\textsc{TopK-BERT} & 30s & 37.24 & 69.36 & 64.65 & 66.79 & 51.30 & 52.13 & 56.91 \\
\textsc{EPR} & 30s & 42.82 & 75.98 & 80.76 & 68.57 & 56.82 & 64.20 & 64.86 \\
\textsc{CEIL} ($n$=50) & 30s & 45.78 & 81.37 & 84.37 & 71.79 & 57.84 & 66.62 & 67.96 \\
\textsc{CEIL} ($n$=100) & 36s & 47.05 & 80.15 & 85.41 & \textbf{73.21} & 59.91 & 67.43 & 68.86 \\
\textsc{CEIL} ($n$=200) & 55s & 46.59 & 80.88 & 85.21 & \textbf{73.21} & 60.26 & 67.15 & 68.88 \\
\textsc{CEIL} ($n$=400) & 87s & 47.14 & \textbf{82.11} & 85.46 & 72.86 & \textbf{60.59} & \textbf{67.52} & 69.28 \\
\textsc{CEIL} ($n$=800) & 118s & \textbf{47.32} & 81.86 & \textbf{86.21} & 72.86 & 60.26 & 67.43 & \textbf{69.32} \\
\bottomrule
\end{tabular}}
\end{table*}

\paragraph{SMCalFlow~\citep{andreas2020task, yin2021compositional}} is a large dialogue dataset, featuring natural conversations about tasks involving calendars, weather, places, and people. The meaning representation is an executable dataflow program featuring API calls, function composition, and complex constraints. 
The SMCalFlow-CS~\citep{yin2021compositional} dataset is a subset of SMCalFlow, containing single-turn natural sentences involving two domains (organization structure and event creation), each having its own set of program symbols. The cross-domain (C) test set evaluates examples that incorporate compositional abilities, while the single-domain (S) test set contains examples from a single domain. On few-shot settings (split $k$-C, where $k\in \{8, 16, 32\}$), the training set includes additional $k$ cross-domain examples, which provide composition symbols, in the evaluation.

% \begin{figure*}[t]
% \centering
% \captionof{table}{}
% \includegraphics[width=6.5in]{figs/examples.pdf}
% \label{fig:examples}
% \end{figure*}

\subsection{Experimental Setup for Compositionality}
We include all the few-shot examples in the context to provide compositional symbols, and we retrieve single-domain exemplars with different retrievers.
We omit the evaluation on 16-C and 32-C splits for the GPT-Neo model as we have no extra room due to the restriction of the input length. 
On Codex, we limit the number of in-context examples to 16 to fairly compare results across the different $k$-C splits.

\section{Additional Experiments}

\subsection{Varying $n$ at Inference Time}

As discussed in \S\ref{sec:inference} that we arrow down the candidate
space with KNN retriever at inference time, we further conducted experiments on multiple datasets to investigate the effect of varying $n$. We show the inference latency on SST-5 validation set and evaluation metrics on different datasets in Table~\ref{tab:vary-n}.
Overall, we found that increasing $n$ tends to improve performance, indicating that increasing $n$ provides a larger exploration space and a higher chance of finding a better subset. 
In addition, inference efficiency is also an important consideration. The latency on the SST5 validation set demonstrates that increasing $n$ will add extra overhead due to the complexity of the MAP inference algorithm, which results in a trade-off between performance and efficiency.

\begin{figure*}[ht]
\centering
\includegraphics[width=6in]{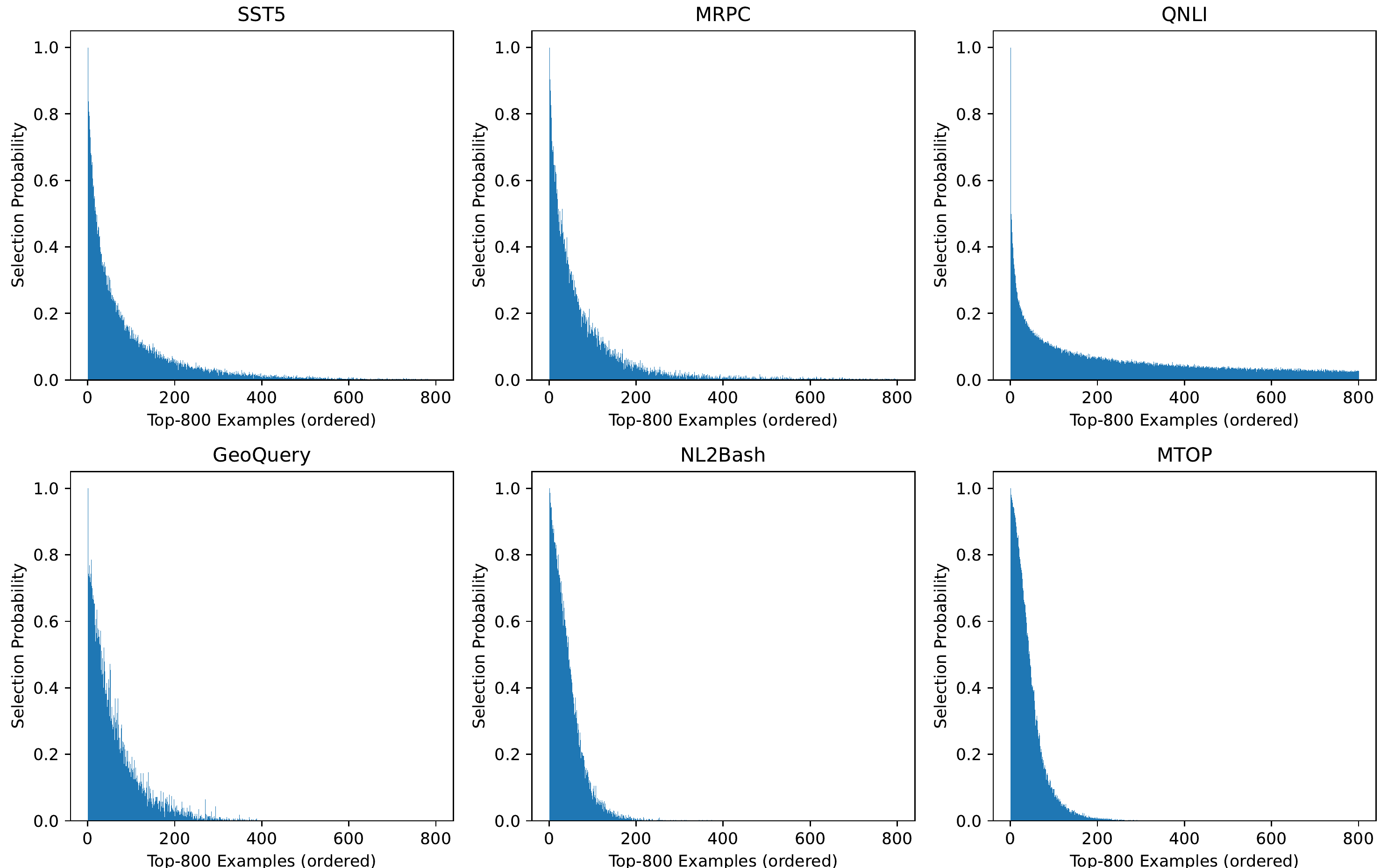}
\caption{Distribution of the selection probability of the top-800 examples. As $n$ increases, its impact on performance diminishes because examples beyond the top 200 are not typically selected on most datasets.}
\label{fig:vary-n}
\end{figure*}

Furthermore, the impact of $n$ on performance tends to become smaller as 
 $n$ increases. We show the distribution of the samples selected in the MAP subset from the top 800 candidate samples in Figure~\ref{fig:vary-n}. Since both relevance and diversity are considered but relevance tends to have greater weight, the impact of $n$ on performance diminishes because examples beyond the top 200 are not typically selected on most datasets.
Therefore, although theoretically, a larger $n$ will have a greater chance of finding a subset, from the perspective of the performance-efficiency trade-off and the diminishing returns of increasing $n$, we adopted an approximate approach that chooses a moderate amount of $n$.

\subsection{Number of In-context Examples}
\label{app:num-ice}
\begin{figure*}[t]
\centering
\includegraphics[width=6in]{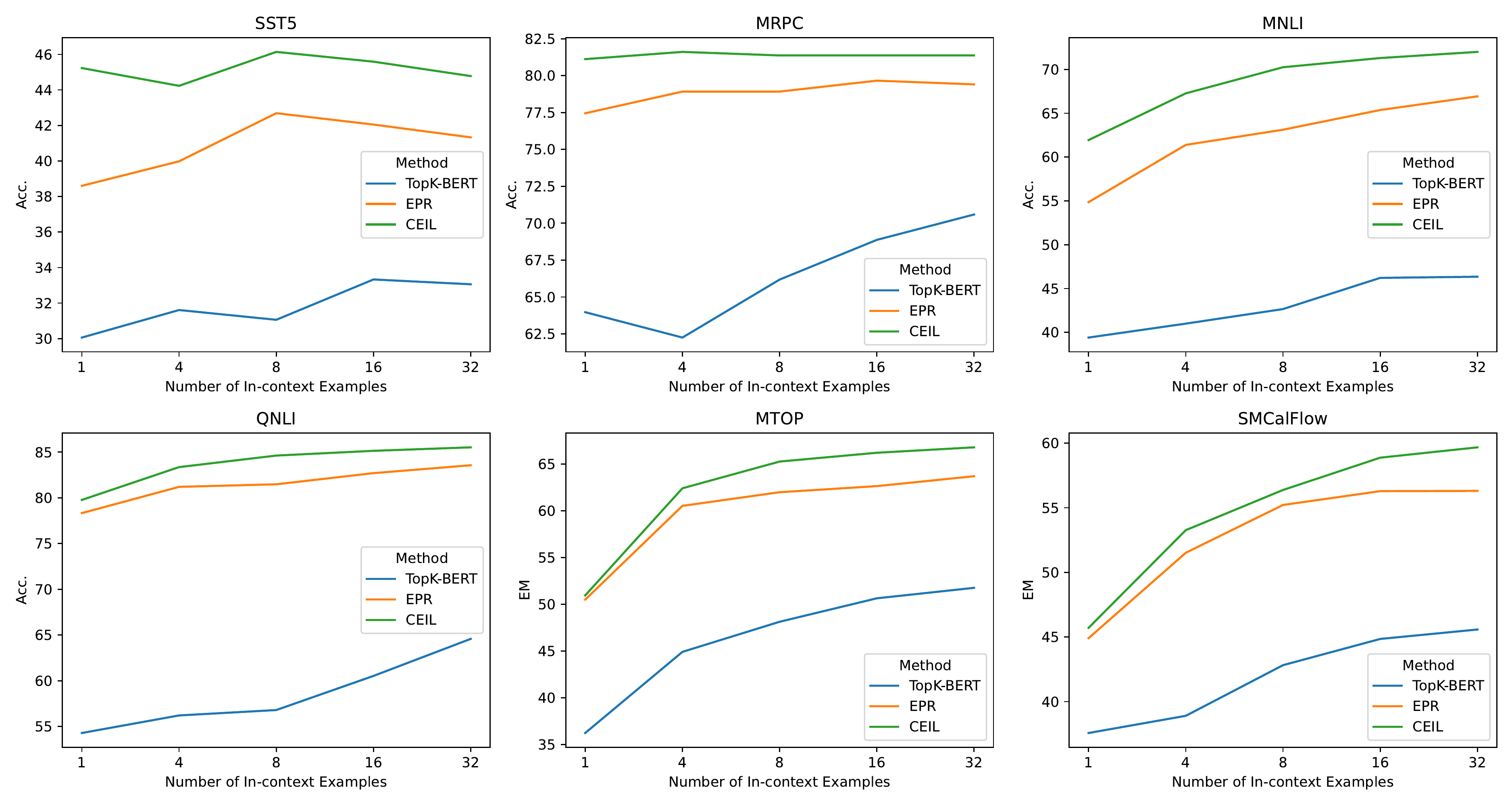}
\caption{Comparison with baselines under various numbers of in-context examples.}
\label{fig:num-ice}
\end{figure*}

We show additional results on the effect of in-context examples in Figure~\ref{fig:num-ice}. We find \ourmodel mostly outperforms \textsc{EPR} and \textsc{TopK-BERT} with 32 in-context examples by using merely 4 and 1 example, respective, greatly cutting down the computations as the attention module ~\citep{vaswani2017attention} in most LMs is of quadratic complexity.

\section{Limitation}
The main limitation of \ourmodel is inherent in the learning-based approach, which performs significantly better than learning-free methods but requires a certain amount of data to train the retriever for each task. The scoring stage in dataset construction of \ourmodel is also slower than EPR since we have to put an in-context example subset into the context instead of a single example. 
Although we have explored the transferability of the retriever, this research is still in its early stages. One potential avenue for future research is to use multitask-tuning to train a unified retriever so that the retriever can be applied directly to new tasks like in the learning-free approaches, without the need to retrain the retriever with new task data.

\end{document}